\documentclass[draftcls,final, onecolumn]{IEEEtran}

\ifCLASSINFOpdf
\else
   \usepackage[dvips]{graphicx}
\fi
\usepackage{url}

\usepackage{graphicx}
\usepackage{amsmath}
\usepackage{amssymb}
\usepackage{amsthm}

\newcommand{\prob}[1]{\mathbb{P}_{#1}}
\newcommand{\probi}[1]{\mathbb{P}_X^{(#1)}}
\newcommand{\Lfunc}{L(\prob{X}, \prob{g(Z)})}
\newcommand{\diam}[1]{\mathrm{diam}(\mathcal{#1})}
\newcommand{\diami}[1]{\mathrm{diam}(\mathcal{X}_{#1})}

\newcommand{\eu}[3]{\big|\big| #1-#2 \big|\big|_{#3}^2}
\newcommand{\eufull}[3]{||[#3^{-1/2}] (#1-#2)||^2}
\newcommand{\g}{\mathcal{G}}
\newtheorem{lemma}{Lemma}[section]

\begin{document}

\title{Limits of Deepfake Detection: A Robust Estimation Viewpoint}

\author{Sakshi Agarwal and Lav R. Varshney, \IEEEmembership{Senior Member, IEEE}\thanks{This work was supported in part by NSF grant CCF-1717530 and is based in part on a thesis submitted in partial fulfillment of the requirements for the degree of Master of Science in the Department of Electrical and Computer Engineering at the University of Illinois at Urbana-Champaign.}
\thanks{S.~Agarwal and L.~R.~Varshney are with the Department of Electrical and Computer Engineering and the Coordinated Science Laboratory, University of Illinois at Urbana-Champaign, Urbana, IL 61801, USA (e-mail: \{sakshi2, varshney\}@illinois.edu).}
}
\maketitle

\begin{abstract}
Deepfake detection is formulated as a hypothesis testing problem to classify an image as genuine or GAN-generated. A robust statistics view of GANs is considered to bound the error probability for various GAN implementations in terms of their performance. The bounds are further simplified using a Euclidean approximation for the low error regime. Lastly, relationships between error probability and epidemic thresholds for spreading processes in networks are established. 
\end{abstract}


\IEEEpeerreviewmaketitle

\section{Introduction}
 
\textit{Deepfake}, a portmanteau of ``deep learning'' and ``fake'' refers to realistic audiovisual content artificially generated using advanced generative algorithms like generative adversarial networks (GANs), with an implied use for unethical purposes. Concerns have been raised over the dissemination of misinformation via online channels \cite{day2019}, and the mistrust that deepfakes engender \cite{guardian}. 

 The gravity of the deepfake problem has been outlined in \cite{deepfake_rep}. Technologies capable of generating hyperrealistic fake images and videos are used for dating scams, ``astroturfing,'' ``catfishing,'' and to gain the victim's trust for blackmail, harassment, or sabotage \cite{catfishing}. With an active community of developers creating free and easy tools to commoditize such technology, there is eroded trust in visual content \cite{courts}. Such ``democratization of fraud,'' can have particularly grave consequences for politics and international affairs where it could be used to incite violence, discredit leaders and institutions, or even tip elections \cite{post_truth_war} and exacerbate disinformation wars that disrupt domestic politics \cite{Gabon, india, nytimes}. 

Such alarming consequences of deepfakes present an urgent need to computationally discern fake content from genuine as they are almost indistinguishable to humans. While several studies use features extracted based on visual artifacts, image quality, lipsync, blinking, or warping for classification (and might soon be obsolete)  \cite{idiap_report, visualartifacts,blinking, warping}, this work gives a generalizable statistical framework with guarantees on its reliability. In particular, we build on the information-theoretic study of authentication \cite{maurer2000} to cast deepfake detection as a hypothesis testing problem specifically for outputs of GANs, themselves viewed through a generalized robust statistics framework \cite{gao2019, Gans}.

\section{Problem Formulation}
\subsection{GAN Formulation}

The GAN trains on infinite sample images with distribution $\prob{X}$, to select an optimal generator function $\hat{g}$ from a family of functions $\mathcal{G}$, which takes an input of Gaussian noise $Z \sim \mathcal{N}(0,I_r)$ to generate an output distribution $\prob{\hat{g}(Z)}$ closest to the input distribution as follows: 
\begin{equation} 
\label{eq:g_hat}
    \hat{g}= \arg \min_{g\in \mathcal{G}} \Lfunc .
\end{equation}
Here $L : \mathcal{M}(\mathbb{R}^d) \times \mathcal{M}(\mathbb{R}^d) \rightarrow \mathbb{R}_{\geq 0}$ measures the distance between the two distributions. We assume that the minimum is attained. The alternate case can be dealt with by a standard limiting argument. 

\subsubsection{Perturbation View of GAN }
Intuitively, the generator family $\mathcal{G}$ of the GAN should be designed such that $\inf_{g\in \mathcal{G}} \Lfunc $ is small. We view GANs as a generalization of the robust statistics framework \cite{Gans} with the true distribution $\prob{X}$ as a slightly perturbed version of a generated distribution $\prob{g(Z)}$ under the distance measure $L(\cdot, \cdot)$. Define: 
\begin{equation}
\label{eq:OPT_def}
    OPT:= \inf_{g\in \mathcal{G}} \Lfunc  
\end{equation}
as the oracle error, which is the minimum distance between the generated and input distributions, for a particular GAN and input distribution. The oracle error is fixed for a choice of $L$ and $\mathcal{G}$. Approximately solving the minimization problem by training on finite samples from $X$, we still get a generated distribution within $O(OPT)$ distance of the true input distribution \cite{Gans}. For error analysis, we choose the following $L$ functions --- Kullback-Leibler (KL) divergence, total variational distance, Jensen-Shannon divergence \cite{goodfellow}, and Wasserstein metric \cite{wgan}. 

\subsection{Hypothesis Test}
The distribution of legitimate images is $\prob{X}$. The distribution generated by the GAN is $\prob{g(Z)}$. The $n$ i.i.d.\ pixels of the input image are $Y_1, Y_2, Y_3,\ldots,Y_n$. Accordingly, the test is
\begin{align*}
    H_0 &:= Y\sim \prob{X} & \text{---Legitimate}  \\
    H_1 &:= Y\sim \prob{{\hat{g}}(Z)} & \text{---GAN generated}
\end{align*}
where $\hat{g}$ is close to the optimal result defined in \eqref{eq:g_hat}. 
The assumption of i.i.d.\ pixels is to make the hypothesis test easier than the non i.i.d.\ case seen in realistic images.

\section{Error Bounds}
\subsection{General Hypothesis Testing Bounds}
We have the following error bounds \cite[ch.~11]{coverthomas} for hypothesis testing:
\begin{align}
 &\text{Neyman-Pearson error:} \ \beta_n^\epsilon \dot{=}\exp( -n D(\prob{X} || \prob{\hat {g}(Z)}) ) \\
 &\text{Bayesian error:} \ P_e^{(n)} \leq \exp \{ -n C(\prob{X},\prob{\hat {g}(Z)}) \}.
\end{align}
where $C$ is the Chernoff information, which can further be bounded in terms of total variational distance \cite{cite_chernoffinfo}.
\begin{equation}
\label{Cbound}
   C(\prob{X},\prob{\hat {g}(Z)}) \geq - \frac{1}{2}\log(1- TV(\prob{X},\prob{\hat {g}(Z)})^2)
\end{equation}
We assume $n$ is sufficiently large for asymptotic bounds to be valid, which mostly holds as the images we generally deal with are fairly high resolution (approximately $10^6$).

\subsection{Deriving Bounds for Specific $L$-functions}
We derive bounds for the Neyman-Pearson and Bayesian error probabilities for the chosen $L$-function implementations of the GAN. 

\subsubsection{KL Divergence}
We have
\begin{equation*}
    \Lfunc= D(\prob{X} || \prob{g(Z)}) \geq OPT \hspace{1cm} \forall g\in \mathcal{G}.
\end{equation*}

\textbf{Neyman-Pearson:}
\begin{align}
    \beta_n^\epsilon \dot{=}\exp( -n D(\prob{X} || \prob{\hat {g}(Z)}) ) 
    \leq \exp(-n OPT).  
\end{align}

\textbf{Bayesian:} From the bound on Chernoff information in \eqref{Cbound}
\begin{align}
    P_e^{(n)} &\leq \exp \{ \frac{n}{2} \log(1- TV(\prob{X},\prob{\hat {g}(Z)})^2) \} 
    \intertext{\quad and from reverse Pinsker's inequality \eqref{rev_pinsker}, assuming $P_g^* = \min_x \prob{\hat {g}(Z)}(x)>0$,}
    P_e^{(n)} &\leq \exp \{\frac{n}{2} \log (1-\frac{P_g^*}{4}D(\prob{X}||\prob{\hat {g}(Z)})) \} \\
    &\leq  \exp \{\frac{n}{2} \log (1-\frac{P_g^*}{4}OPT )\}  \\
    &= \left(1-\frac{P_g^*}{4}OPT \right)^{n/2}.
\end{align}

\subsubsection{Total Variational Distance}
We have 
\begin{equation*}
    \Lfunc= TV(\prob{X}, \prob{g(Z)}) \geq OPT \hspace{1cm} \forall g\in \mathcal{G}.
\end{equation*}

\textbf{Neyman-Pearson:} Using Pinsker's inequality \eqref{pinsker}
\begin{align}
    \beta_n^\epsilon 
    &\leq \exp(-2 n TV(\prob{X} || \prob{\hat {g}(Z)})^2 ) \\
    &\leq \exp(-2 n OPT^2). 
\end{align}

\textbf{Bayesian:} From the bound on Chernoff information in \eqref{Cbound}
\begin{align}
    P_e^{(n)}     &\leq \exp \{ \frac{n}{2} \log(1- TV(\prob{X},\prob{\hat {g}(Z)})^2) \} \\
    &\leq \exp \{ \frac{n}{2} \log(1- OPT^2)  \\
    &=(1- OPT^2)^{n/2}. 
\end{align}
\subsubsection{Jensen-Shannon Divergence}
From the bound on Jensen-Shannon divergence \eqref{JSbound}:
\begin{equation}
    2TV(\prob{X}, \prob{g(Z)}) \geq JS(\prob{X}, \prob{g(Z)}).
\end{equation}
Thus, we have $\forall g \in \g $
\begin{equation}
\label{eq:JSOPT}
        2TV(\prob{X}, \prob{g(Z)}) \geq JS(\prob{X}, \prob{g(Z)})  \geq OPT.
\end{equation}

\textbf{Neyman-Pearson:}
From Pinsker's inequality and \eqref{eq:JSOPT}
\begin{align}
    \beta_n^\epsilon  &\leq \exp(-2 n TV(\prob{X} || \prob{\hat {g}(Z)})^2 ) \\
    &\leq \exp (-\frac{n}{2} OPT^2)
\end{align}

\textbf{Bayesian:} From the bound on Chernoff information in \eqref{Cbound}
\begin{align}
    P_e^{(n)} 
    &\leq \exp \{ \frac{n}{2} \log(1- TV(\prob{X},\prob{\hat {g}(Z)})^2) \} \\
    &\leq \exp \left\{ \frac{n}{2} \log\left(1- \frac{OPT^2}{4}\right) \right\} \\
   &= \left(1- \frac{OPT^2}{4}\right)^{n/2}.
\end{align}

\subsubsection{Wasserstein Metric}
From the bound on Wasserstein metric \eqref{Wbound} 
\begin{equation}
\label{eq:WOPT}
    \diam{X}TV(\prob{X}, \prob{g(Z)}) \geq W(\prob{X}, \prob{g(Z)}) \geq OPT.
\end{equation}
where $\diam{X}$ is the diameter of the space.

\textbf{Neyman-Pearson:} From Pinsker's inequality and \eqref{eq:WOPT}
\begin{align}
    \beta_n^\epsilon &\leq \exp(-2 n \ TV(\prob{X} || \prob{\hat {g}(Z)})^2 ) \\
    &\leq  \exp \left(-\frac{2n}{\diam{X}^2}OPT^2 \right)
\end{align}

\textbf{Bayesian:} From the bound on Chernoff information in \eqref{Cbound}
\begin{align}
    P_e^{(n)} 
    &\leq \exp \{ \frac{n}{2} \log(1- TV(\prob{X},\prob{\hat {g}(Z)})^2) \} \\
    &\leq \exp \bigg\{ \frac{n}{2} \log\left(1- \left(\frac{OPT}{\diam{X}}\right)^2\right)\bigg\} \\
    &= \left(1- \left(\frac{OPT}{\diam{X}}\right)^2\right)^{n/2} .
\end{align} 

\begin{table}
    \centering
    \caption{Summary of the error bounds}
    \label{table1}
    \begin{tabular}{cll}
    \hline
        L-function & \multicolumn{1}{p{2cm}}{\centering Neyman-Pearson Bound \\ $\beta_n^\epsilon \leq$}    & \multicolumn{1}{p{2cm}}{\centering Bayesian Bound \\ $P_e^{(n)}\leq$}   \\
        \hline
        KL divergence & $ \exp(-n OPT)$ & $(1-\frac{P_g^*}{4}OPT )^{n/2} $\\
        TV distance & $ \exp(-2n OPT^2)$ & $(1-OPT^2)^{n/2}$\\
        JS divergence & $\exp(-\frac{n}{2} OPT^2)$ &$(1-\frac{1}{4}OPT^2 )^{n/2}$ \\
        Wasserstein & $\exp \left(-\frac{2n}{\diam{X}^2}OPT^2 \right)$ & $\left(1- \left(\frac{OPT}{\diam{X}}\right)^2\right)^{n/2} $  \\
        \hline
    \end{tabular}
\end{table}

\subsection{Discussion}
The bounds are summarized in Table \ref{table1}. As $OPT$ increases, i.e., the GAN used is less accurate, it is exponentially easier to detect deepfakes in the Neyman-Pearson case, and polynomially so in the Bayesian case.  The bound decays exponentially with the resolution $n$. Thus, if we require very high resolution in order to trust images, an extremely accurate GAN would be required to go undetected.

\subsection{Generalizing the Problem}
Images typically have regions depicting features like eyes, mouth, and hair or objects like vehicles and trees that follow a common distribution. Consider the image to consist of $k$ patches with $m$ pixels each, following the distribution $\probi{i}$, such that $m$ is still large for the hypothesis testing bounds to be valid. We now similarly define individual oracle errors for each patch as $OPT_i:= \inf_{g\in \mathcal{G}}L(\probi{i}, \prob{g(Z)})$, considering that the GAN generates each patch independently. The generalized bounds can be obtained \emph{mutatis mutandis} by multiplying together the corresponding bounds for each of the $k$ patches obtained by using $m$ for $n$, $OPT_i$ for $OPT$, $P_{g_i}^*$ for $P_g^*$, and $\diami{i}$ for $\diam{X}$ to represent the values defined for each patch of the image.

\section{Euclidean Information Theory}
With advances in deep learning and cheap availability of data, GANs have grown ever more complex and can generate a very close version of the true distribution. This development suggests further analysis in the low error regime using the concept of Euclidean information theory, which gives the following approximation for KL divergence \cite{euclidean}. 

\begin{lemma}[Euclidean approximation]
 \label{euclidean_lemma}
 When two distributions are close to each other, i.e. $P\approx Q$ , the KL divergence between them can be approximated as 
 \begin{equation}
     D(P||Q) \approx \frac{1}{2}\eu{Q}{P}{\hat{P}}  = \frac{1}{2} \eufull{Q}{P}{\hat{P}}
 \end{equation}
 where the weight $\hat{P}$ is any distribution in the neighborhood of P and Q. 
 \end{lemma}

 \subsection{Simplified Bounds Using Euclidean Approximation}
On applying the approximation in the hypothesis testing error bounds, we get the following:
 \begin{equation}
     \beta_n^{\epsilon} \dot{=} \exp (-nD(\prob{X}||\prob{\hat{g}(Z)})) \approx \exp(-\frac{n}{2} \eu{\prob{X}}{\prob{\hat{g}(Z)}}{P})
 \end{equation}
where the weight (subscript $P$) is any distribution in the neighborhood of the two distributions. Also, from \cite[ch.~11]{coverthomas}
\begin{gather}
    P_e^{(n)}\leq \exp(-nD^*(\prob{X}||\prob{\hat{g}(Z)})), \  \text{where}\\
    D^*(\prob{X}||\prob{\hat{g}(Z)}) = \max_\lambda \min (D(\prob{X}|| \prob{\lambda}), D(\prob{\hat{g(Z)}}||\prob{\lambda})) \\
    \prob{\lambda}= \frac{\prob{X}^\lambda \prob{\hat{g}(Z)}^{1-\lambda}}{\sum_{a\in \mathcal{X}}  \prob{X}^\lambda(a) \prob{\hat{g}(Z)}^{1-\lambda}(a)}.
\end{gather} 
    Assuming $\prob{X}\approx \prob{\hat{g}(Z)}, \prob{\lambda}$ is also close to the two distributions. So using the approximation in Lemma \ref{euclidean_lemma}, the maximizing $\lambda$ is such that the two quantities are equal. 
\begin{align}
       D^*(\prob{X}||\prob{\hat{g}(Z)})  
    &\approx \frac{1}{2} \eu{\frac{\prob{X}+\prob{\hat{g}(Z)}}{2}}{\prob{X}}{\prob{X}} \\ 
    &= \frac{1}{8} \eu{\prob{X}}{\prob{\hat{g}(Z)}}{\prob{X}} \\
    \label{eq:c_approx}
    &\approx \frac{1}{4} D(\prob{X}||\prob{\hat{g}(Z)}).
\end{align}

 \subsubsection{KL Divergence}
 \textbf{Neyman-Pearson:} The bound stays the same. 

\textbf{Bayesian:}
\begin{align}
P_e^{(n)} 
&\leq \exp \big(-\frac{n}{4} D(\prob{X}||\prob{\hat{g}(Z)}) \big) \\
&\leq \exp \big( -\frac{n}{4} OPT \big).
\end{align}

\subsubsection{Total Variational Distance}
\textbf{Neyman-Pearson:} The bound stays the same. 

\textbf{Bayesian:} Using Pinsker's inequality
\begin{align}
P_e^{(n)} &\leq \exp \big(-\frac{n}{4} D(\prob{X}||\prob{\hat{g}(Z)}) \big) \\
&\leq \exp \big( -\frac{n}{4} 2TV(\prob{X}||\prob{\hat{g}(Z)})^2\big) \\
&\leq \exp \big(-\frac{n}{2} OPT^2 \big) 
\end{align}

\subsubsection{Jensen-Shannon Divergence}
When $P_1 \approx P_2$ , Jensen-Shannon divergence can also be approximated in the Euclidean regime as:
\begin{align}
    JS(P_1, P_2) &= \frac{1}{2}[D(P_1||\tfrac{P_1+P_2}{2}) + D(P_2||\tfrac{P_1+P_2}{2})]\\
    &\approx \frac{\eu{P_1}{P_2}{P_1}}{4}  
    \approx \frac{D(P_1 || P_2)}{2}.
\end{align}
Thus, we have 
\begin{equation}
    OPT \leq JS(\prob{X}, \prob{g(Z)}) \approx \frac{D(\prob{X} || \prob{g(Z)})}{2}.
\end{equation}

\textbf{Neyman-Pearson:}
\begin{align}
    \beta_n^\epsilon &\dot{=}\exp( -n D(\prob{X} || \prob{\hat {g}(Z)}) ) 
    \leq \exp (-2n \ OPT)
\end{align}

\textbf{Bayesian:}
\begin{align}
P_e^{(n)} &\leq \exp \big(-\frac{n}{4} D(\prob{X}||\prob{\hat{g}(Z)}) \big) 
\leq \exp \big(-\frac{n}{2} \ OPT) \big) 
\end{align}

\subsubsection{Wasserstein Metric}
Combining the bound for Wasserstein metric in \eqref{Wbound} and Pinsker's inequality, we have
\begin{equation}
OPT \leq W(\prob{X}, \prob{g(Z)})
\leq \diam{X} \sqrt{\frac{D(\prob{X}, \prob{g(Z)})}{2}}.
\end{equation}

\textbf{Neyman-Pearson:} The bound remains the same. 

\textbf{Bayesian:}
\begin{align}
P_e^{(n)} &\leq \exp \big(-\frac{n}{4} D(\prob{X}||\prob{\hat{g}(Z)}) \big) \\
&\leq\exp  \left(-\frac{n}{2 \ \diam{X}^2} OPT^2 \right) 
\end{align}

\begin{table}
    \centering
    \caption{Simplified bounds using Euclidean approximation}
    \label{table2}
    \begin{tabular}{cll}
    \hline
        L-function & \multicolumn{1}{p{2cm}}{\centering Neyman-Pearson Bound \\ $\beta_n^\epsilon \leq$}    & \multicolumn{1}{p{2cm}}{\centering Bayesian Bound \\ $P_e^{(n)}\leq$}   \\
        \hline
        KL divergence & $ \exp(-n OPT)$ & $\exp(-\frac{n}{4} OPT) $\\
        TV distance & $ \exp(-2n OPT^2)$ & $\exp(-\frac{n}{2} OPT^2)$\\
        JS divergence & $\exp(-2n OPT)$ &$\exp(-\frac{n}{2} OPT)$ \\
        Wasserstein & $\exp \left(-\frac{2n}{\diam{X}^2}OPT^2 \right)$ & $\exp \left(-\frac{n}{2\ \diam{X}^2}OPT^2 \right) $  \\
    \hline
    \end{tabular}
\end{table}

\subsection{Discussion}
Table \ref{table2} summarizes the simplified bounds using Euclidean approximation. For the same value of $OPT$, we see that GANs with Jensen-Shannon divergence as the $L$-function are the easiest to detect with lowest bound on the Bayesian error probability. Also, while the relationship between the error probability and image resolution remains the same, it is exponentially easier to detect deepfakes from GANs with higher $OPT$ for both tests unlike the results in Table \ref{table1}. The exponent of Neyman-Pearson bound is simply that of the Bayesian bound multiplied by a factor of four for all $L$-functions.

\section{Epidemic Threshold Theory and Deepfakes}
Deepfakes may disperse rapidly in social networks \cite{fb_slam} with spreading dynamics similar to diseases. The SIR (susceptible-infected-recovered) model for epidemics can be useful to assess the risk posed by a deepfake. Epidemic threshold characterizes the critical level $\lambda_c$ for effective spreading rate $\lambda$ above which a global epidemic occurs and the spreading cannot be contained. The threshold can be predicted based purely on the network structure \cite{SIR_model}. The spreading rate is expressed as 
\begin{equation}
    \lambda= \frac{\beta}{\gamma}
\end{equation}
where $\beta$ is the probability of transmission from an infected to susceptible node, and $\gamma$ is the probability of recovery, i.e., probability of correctly detecting the deepfake $1-P_e$. Let the bound for $P_e^{(n)}$ be $\exp (-n f(OPT))$. Thus, we have
\begin{equation}
    \lambda \leq \frac{ \beta}{1- \exp(-n  f(OPT) )}.
\end{equation}
 Along with the condition for containment $\lambda \leq \lambda_c$, we get the following condition on $OPT$ for the deepfake to be locally containable.
\begin{equation}
      f(OPT) \geq -\frac{1}{n} \ln   \left(1- \frac{ \beta}{\lambda_c} \right).
\end{equation}
When $f(OPT)$ is greater than the expression on the right, which depends only on $n$ and the network structure, the deepfake can be locally confined. 
From Tables \ref{table1} and \ref{table2}, $f$ is always increasing. A higher $OPT$, i.e., worse GAN accuracy, guarantees higher robustness to global spread of misinformation. Thus the network will be less easily fooled by a weaker deepfake generation system.

\section{Conclusion}
 This work provides a statistical framework to detect deepfakes and error guarantees for these tests. Avenues to extend this study include--- accounting for a possible gender, race, etc.\ based bias in the training data to prevent incorrect classification, developing a similar framework for conditional GANs \cite{condGAN} for detecting face swapped content, and deriving error bounds in terms of more commonly used GAN evaluation metrics as in \cite{gan_metrics}. Further, by incorporating these results with studies on spreading dynamics of infectious diseases  \cite{bayesian_network,SIR_source_detection}, we can obtain bounds on source detection probability for deepfakes. 

\bibliographystyle{IEEEtran}
\bibliography{files/ref}

\appendix
\renewcommand{\theequation}{A.\arabic{equation}}
\section*{Important Inequalities}

\subsection{Pinsker's inequality:}
\begin{equation}
\label{pinsker}
    TV(P_1, P_2)^2 \leq  \frac{1}{2} D(P_1|| P_2).
\end{equation}
\begin{proof}
The proof is given in \cite{f-divergence}. 
\end{proof}

\subsection{Reverse Pinsker's inequality:}
For the case of finite alphabet $\mathcal{X}$, when $P_2^*= \min_{x\in\mathcal{X}} P_2(x) > 0 $, 
\begin{equation}
\label{rev_pinsker}
    \frac{4}{P_2^*}TV(P_1, P_2)^2 \geq  D(P_1|| P_2).
\end{equation}
\begin{proof}
This reverse bound on total variational distance has been proved by Csisz{\'a}r and Talata \cite[p.~1012]{cite_reversepinsker}.
\end{proof}

\subsection{Bound on Jensen-Shannon divergence:}
\begin{equation}
\label{JSbound}
    JS(P_1, P_2) \leq 2 TV(P_1, P_2).
\end{equation}
\begin{proof}
The proof is given in \cite{cite_JSbound}.
\end{proof}

\subsection{Bound on Wasserstein metric:}
\begin{equation}
\label{Wbound}
    W(P_1, P_2) \leq \diam{X} TV(P_1, P_2)
\end{equation}
where $\diam{X}= \sup \{d(x,y) : (x,y)\in \mathcal{X} \} $ is the diameter of the space. 
\begin{proof}
The proof is given in  \cite{cite_Wassersteinbound}.
\end{proof}

\end{document}